\documentclass[10pt]{article} % For LaTeX2e
\usepackage[accepted]{tmlr}
\usepackage{amsmath,amsfonts,bm}

\def\eqref#1{equation~\ref{#1}}
\def\1{\bm{1}}

\DeclareMathAlphabet{\mathsfit}{\encodingdefault}{\sfdefault}{m}{sl}
\SetMathAlphabet{\mathsfit}{bold}{\encodingdefault}{\sfdefault}{bx}{n}

\usepackage{hyperref}
\usepackage{url}

\usepackage{xspace}
\usepackage{amsmath}
\usepackage{amssymb} 
\usepackage{amsfonts}
\usepackage{graphicx}
\usepackage{subcaption}
\usepackage{booktabs, tabularx}
\usepackage{multirow} 
\usepackage{array}
\usepackage{wrapfig}
\usepackage{algorithm}
\usepackage{algpseudocode}

\newcommand{\framework}{\textsc{MemLLM}\xspace}
\newcommand{\docred}{DocRED\xspace}

\usepackage[most]{tcolorbox}
\definecolor{my_gray}{RGB}{239,239,239}
\definecolor{my_lyellow}{RGB}{251,230,163}
\definecolor{my_dyellow}{RGB}{234,196,82}
\definecolor{my_blue}{RGB}{211,226,241}
\definecolor{my_green}{RGB}{220,234,213}

\long\def\devour#1{\ignorespaces} 

\usepackage{xcolor}
\usepackage{soul}

\newcommand{\hlc}[2][yellow]{{%
    \colorlet{foo}{#1}%
    \sethlcolor{foo}\hl{#2}}%
}

\def\mathlinebreak{\\[0.1cm]}
\def\mathindent{\mbox{\hspace{0.5cm}}}

\usepackage{tikz}
\usepackage{etoolbox}
\newcommand{\circled}[2][]{%
  \tikz[baseline=(char.base)]{%
    \node[shape = circle, draw, inner sep = 1pt]
    (char) {\phantom{\ifblank{#1}{\small #2}{\small #1}}};%
    \node at (char.center) {\makebox[0pt][c]{\footnotesize #2}};}}
\robustify{\circled}

\title{\framework: Finetuning LLMs to Use Explicit Read-Write Memory}
\author{\name Ali Modarressi$^{1,2}$, Abdullatif Köksal$^{1,2}$, Ayyoob Imani$^{1,2}$, Mohsen Fayyaz$^{3}$, Hinrich Schütze$^{1,2}$ \\\email amodaresi@cis.lmu.de \\\\
      \addr $^{1}$Center for Information and Language Processing, LMU Munich, Germany \\
      $^{2}$Munich Center for Machine Learning, Germany \\
      $^{3}$Microsoft, Berlin, Germany}

\def\month{04}  % Insert correct month for camera-ready version
\def\year{2025} % Insert correct year for camera-ready version
\def\openreview{\url{https://openreview.net/forum?id=dghM7sOudh}} % Insert correct link to OpenReview for camera-ready version

\begin{document}

\maketitle

\begin{abstract}
While current large language models (LLMs) perform well on many
knowledge-related tasks, they are limited
by relying on their parameters as an implicit storage
mechanism.  As a result, they struggle with memorizing rare
events and with updating their memory as facts change over
time.  In addition, the uninterpretable nature of parametric
memory makes it challenging to prevent hallucination.
Model editing and augmenting LLMs with parameters specialized for memory
are only partial solutions. In this paper, we
introduce \framework, a novel method of enhancing LLMs by
integrating a structured and explicit read-and-write memory
module. \framework tackles the aforementioned challenges by
enabling dynamic interaction with the memory and improving
the LLM's capabilities in using stored knowledge. Our
experiments indicate that \framework enhances the LLM's
performance and interpretability, in language modeling in
general and knowledge-intensive tasks in particular. We
see \framework as an important step towards making LLMs more
grounded and factual through memory augmentation. The project repository is publicly available at: \url{https://github.com/amodaresi/MemLLM}
\end{abstract}

\section{Introduction}

State-of-the-art large language models (LLMs)
perform well in knowledge-intensive
tasks \citep{yu2023generate, ChowdheryPaLM2023}.
%Owing to
%their extensive scale in terms of training data and model
%size,
They solve these tasks utilizing the
information memorized in their vast array of parameters \citep{roberts-etal-2020-much}. However, the
effectiveness of 
parameter-based memorization is limited for
infrequent entities and
concepts \citep{kandpal2023long-tail,mallen-etal-2023-trust} and
is prone to temporal
degradation \citep{kasai2023realtime,jang-etal-2022-temporalwiki}. 
Parametric model editing  may address some of
these
issues \citep{Sinitsin2020Editable,de-cao-etal-2021-editing,pmlr-v162-mitchell22a},
but struggles with maintaining
locality -- possibly damaging model
performance in unrelated areas \citep{yao-etal-2023-editing,
gu2024model}. 
Moreover, model editing  often deteriorates  performance
%when applied to sequential editing or batch updates. This is
when applying sequential editing or batch updates. This is 
because it primarily focuses on applying (and evaluating) single edits one-by-one \citep{huang2023transformerpatcher}.
Finally, model editing may struggle to generalize and maintain previous edits when updating multiple facts simultaneously \citep{yao-etal-2023-editing}.

Other parametric solutions, like augmenting LLMs with extra parameters such as memory pools
can preserve knowledge for subsequent
utilization \citep{wang2023augmenting,wang2024memoryllm}.
However, parametric memorization is prone to distortion
and hallucinated nonfactual output.
In addition, parametric mechanisms like memory pools have limited
capacity and lack interpretability
\citep{maynez-etal-2020-faithfulness, ji2023survey}.
% While Retrieval-Augmented Generation (RAG) methods \citep{lewis2020RAG} can provide
% updated facts, their unstructured
% information storage makes fact editing challenging. Editing an atomic fact requires updating all
% related instances, risking contradictions if conflicting facts are
% retrieved together \citep{shi2023large}.

Another approach is to augment LLMs with a non-parametric
memory component that interacts
with the LLM either through natural
language text
or a formalized API \citep{wang2023interactive}. Although
prior work has demonstrated enhanced abilities in extended
dialogs, long-text generation and
question answering \citep{packer2023memgpt, hu2023chatdb,
zhou2023recurrentgpt}, these methods are primarily
prompt-dependent and necessitate customization for each
specific task and model. They also suffer
from the lack of a structured memory. This undermines
interpretability and
interoperability \citep{wang2023interactive}.
While Retrieval-Augmented Generation (RAG) methods
\citep{lewis2020RAG} can provide updated facts,
their unstructured storage complicates fact editing.
Updating an atomic fact requires modifying all related
instances to prevent contradictions when conflicting
facts are retrieved together \citep{shi2023large}.

In this paper, we introduce \framework, an LLM endowed with
an explicit memory component. This architecture has the
following key characteristics.
\begin{itemize}
\item This explicit memory component  has the general advantages
of some of the memory-focused work we discussed: we can \textbf{keep information
accessible indefinitely}, beyond the context window, including
\textbf{infrequent information} that standard LLMs struggle with.
\item 
The LLM has \textbf{both read and write access} to the
explicit memory component, i.e., it can store information in the
memory as it processes text (or interacts with a user) and
retrieve it when it needs it. % (Figure \ref{fig:memllm_vs_plm}).
\item
We adopt \textbf{finetuning} to train the model to access
the explicit memory component through read and write commands.
To this end,
we specify an \textbf{API for read and write access}.
Based on the API specification, we create a
dataset with training examples of API read and write
commands and finetune the LLM on it.  Our
published training dataset can be used to finetune any
language model, endowing it with an explicit memory
component without
requiring architectural changes.
\item 
The memory component has an explicit structured schema, similar to a
database schema. Therefore, it is \textbf{interpretable} and inspectable for humans;
it is \textbf{editable} by humans; it is \textbf{scalable}
since databases have excellent scalability properties; and  it
is \textbf{interoperable} since the contents of the memory
can be exported (e.g., to a different LLM supporting explicit
memory) and contents can be imported from data resources
(e.g., from Wikidata).
\end{itemize}

Our evaluation on Re-\docred \citep{tan-etal-2022-revisiting}
demonstrates that \framework achieves better perplexity
compared to baselines without memory components, with strong
gains on named entities.
We also show that \framework  outperforms
non-memory-based methods on
knowledge editing.

\section{Related work}

\textbf{External memory augmentation.}  Augmenting an LLM
with memory as an external component can enhance its ability
to process larger contexts and maintain reliable records by
storing facts and knowledge. Such components include
databases, knowledge bases and knowledge graphs that LLMs
interact with via natural or formal language
\citep{guu2020REALM,lewis2020RAG,liu-etal-2022-relational,yao2023react,Park2023GenerativeAgents,zhou2023recurrentgpt,schick2023toolformer,hu2023chatdb}. For
instance, Retrieval Augmented Generation (RAG) retrieves
relevant text snippets from large document databases, to
improve factuality \citep{guu2020REALM,lewis2020RAG}.  Other
recent solutions store summarized information from previous
contexts for future retrieval, improving performance in
long-form generation, summarization, question answering and
dialog coherence \citep{Park2023GenerativeAgents,
  zhou2023recurrentgpt, packer2023memgpt, chen2023walking,
  wang2024enhancing}.

% \updated{
% Of this work on external memory augmentation, knowledge
% graph approaches are notable because knowledge graphs can be
% interpreted as memories of triples, similar to our
% memory. However, while there is work on extracting triples
% from text into knowledge graphs
% \citep{DBLP:conf/emnlp/ZhangS24} using LLMs and on LLMs
% ``interactively'' querying knowledge graphs to answer
% questions \citep{Baek2023KnowledgeAugmentedLM}, our approach
% is distinguished by teaching the LLM memory-write and memory-read
% functionality through finetuning. Writing to
% the memory and reading from it becomes an integral part of how the LLM
% processes text, similar to how humans store information in
% memory and retrieve it constantly as they go through
% life. Similar integrated systems have also been considered
% in the knowledge graph literature (e.g., AutoKG
% \citep{DBLP:journals/www/ZhuWCQOYDCZ24}), but to the best of
% our knowledge our system is the first that goes beyond a
% conceptual proposal and is demonstrated to outperform
% baselines in an empirical evaluation.
% }

In general, our framework aligns with external memory methodologies but
stands out with its structured format for storing
information. This explicit memory facilitates large-scale
knowledge editing and makes the model's output generation
process more interpretable.  While similar structured
storage approaches exist, they are often task-specific.
 For example, \citet{hu2023chatdb} introduce ChatDB,
  which takes a database structure as input for a data
  record management task.  This requires prior knowledge of
  the task-specific database schema (e.g., a sales table),
  which must be defined and provided as a prompt to the
  model.  In contrast, our method is designed for generic
language modeling, making it adaptable to a variety of tasks
without extensive prompt engineering.
Knowledge graphs are 
  another structured format
  that can be  interpreted as a memory of triples, similar to our memory.
However, while there is work on extracting triples
from text into knowledge graphs
\citep{DBLP:conf/emnlp/ZhangS24} using LLMs and on LLMs
``interactively'' querying knowledge graphs to answer
questions \citep{Baek2023KnowledgeAugmentedLM}, our approach
is distinguished by teaching the LLM memory-write and memory-read
functionality through finetuning. 
This makes memory an integral component of the model's text
processing
and at the same time enables the model to flexibly handle new and updated knowledge.
% similar to how humans store information in
% memory and retrieve it constantly as they go through
% life.
Similar integrated systems have also been considered
in the knowledge graph literature (e.g., AutoKG
\citep{DBLP:journals/www/ZhuWCQOYDCZ24}), but to the best of
our knowledge, our system is the first that goes beyond a
conceptual proposal and demonstrates empirical success in both language modeling and knowledge editing. Additionally,
our published
training dataset can be used to endow any trainable language
model with explicit memory without requiring architectural
changes.

\textbf{Memory as a state.}
The term \emph{memory} can refer to
recurrent architectures that represent past context with
vectors \citep{Hochreiter1997LSTM,cho-etal-2014-learning-GRU}. Transformer-based
models use similar mechanisms with memory tokens (to transfer
context across segments) and memory pools (to share
information across multiple
contexts) \citep{burtsev2020memory,bulatov2022recurrent,wang2024memoryllm}.
While recent advances use vector- or parameter-based memory
systems for long-range
dependencies \citep{martins-etal-2022-former,wu-etal-2022-memformer,wu2022memorizing,cheng2023lift,wang2023augmenting,he2024camelot},
they are limited by memory vector
capacity \citep{jelassi2024repeat}. In contrast, \framework
has no such architectural limitations and features explicit,
interpretable and editable memory.

\textbf{Knowledge editing.}
The goal of knowledge editing is to apply data-efficient
changes to a model's behavior for a set of edits  while
keeping other knowledge
unaffected \citep{yao-etal-2023-editing, gu2024model,
zhang2024comprehensive}. Meta-learning and locate-then-edit
are two classes of parametric methods that modify model
weights. In meta-learning, a hypernetwork is trained and
applied to the model weights during test
time \citep{de-cao-etal-2021-editing, mitchell2021fast}. In
locate-then-edit, the weights triggered by a knowledge
expression are located and modified
\citep{dai-etal-2022-knowledge,
meng2022locating}. There are
also memory-based methods that do not alter the original
model weights but use an external
memory \citep{gu2024model}. E.g., methods like
SERAC, GRACE and DEFER use retrieval-based memory to fetch
previously given edits and apply them to new
inputs \citep{pmlr-v162-mitchell22a,
hartvigsen2024aging}. In WISE \citep{wang2024wise},
in addition to the LLM, two additional parametric models are trained:
a side
memory and a routing network.
Based on the  query,
the routing network decides which memory to use, the side
memory or the main LLM.
Evaluations show that 
multiple edits at a time (batch editing) or 
successive edits (sequential editing) are challenging tasks
-- but certainly critical for the intended application of
knowledge editing.
 While most methods can handle a few
edits at a time, their performance drops when applying more
 \citep{yao-etal-2023-editing, wang2024wise}.  Due to
the explicit memory structure of \framework, it can handle a
large number of edits while maintaining performance.

\section{Methodology}
Our approach to endowing an LLM with an explicit memory is
finetuning with the standard language modeling
objective. We now present a finetuning regime
that teaches the LLM (1) to extract knowledge from text and
write it to the memory and (2) to read knowledge from the
memory and leverage it for better language modeling.
Following \citet{schick2023toolformer}
and \citet{modarressi2023ret},
we define an API through which
the LLM initiates memory writes and
memory reads.

% i hope i didn't omit anything crucial from this:
%In our framework, the primary goal is to finetune an LLM to
%incorporate memorization capabilities via an interpretable
%external memory component.  Rather than relying solely on
%parameter-based memorization, which is inherently limited,
%lacks interpretability, and compromises generalization, our
%approach leverages an external memory system for storing and
%retrieving knowledge.  To this end,

%We now describe the data structure of the
%explicit memory, the API and how we finetune the LLM to learn
%the API including how we create the training data for
%finetuning.

\subsection{Memory structure}
\label{sec:mem_struct}
\begin{figure*}[t]
    \centering
    {\includegraphics[width=0.93\linewidth, trim=1 1 1 1, clip] {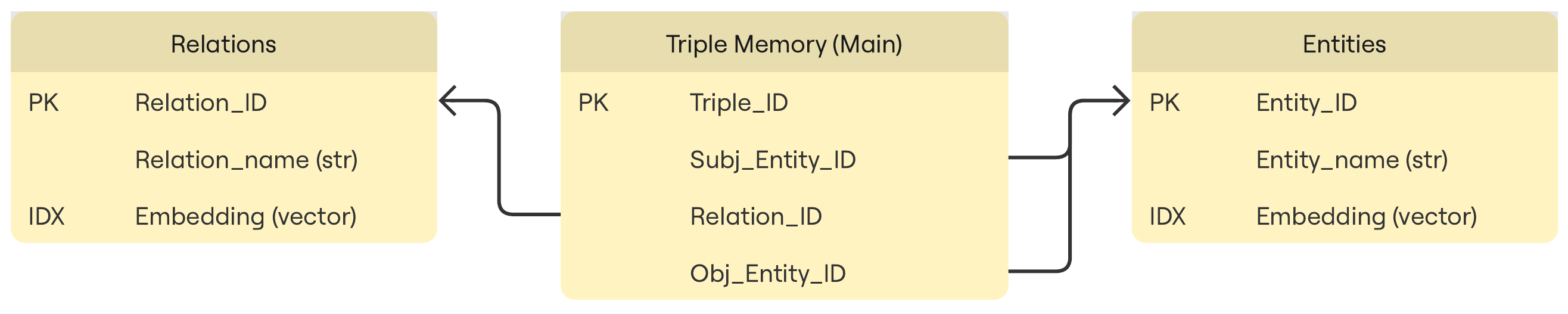}}
    % EDITSTART
    \caption{\framework memory schema. Each triple is stored in the ``Triple Memory'' using its subject entity ID, relation ID and object entity ID, along with a designated triple ID. Entities and relations are stored in separate tables, each containing their designated ID, name and corresponding vector embedding. Both tables are indexed based on their vector embeddings.}
    % EDITEND
    \label{fig:memory_schema}
\end{figure*}

The memory stores information in relation triples.  Each
triple has the form $r = \langle e_s, t, e_o \rangle$, where
$e_s$ is the first entity or subject, $e_o$ the second
entity or object, and $t$ the relation. Example:
$\langle\text{Washington D.C.}, \text{capital of},
\text{United States}\rangle$.  The entities and relations
are stored as raw text and vectors, each in separate tables.
As shown in Figure \ref{fig:memory_schema}, the facts are
stored in the main table ``Triple Memory'' using three
identifiers linked to two other tables: one for subject and
object entities, and one for relations. The entities and
relations tables are indexed by unique names. 
We enforce uniqueness across the three
linked identifiers
in the main
table ``Triple Memory'', i.e., a specific combination
(Entity\_ID$_1$,Relation\_ID,Entity\_ID$_2$) can only occur
once in the table. Because these identifiers indirectly
refer to the names of the entities or relations, different
identifiers imply different entity or relation names,
thereby preventing redundant storage of identical facts.
The vector representations (created with Contriever
\citep{izacard2022unsupervised}) abstract away from
different surface forms of the same entity, e.g., ``US'' vs
``USA''.  In the interest of brevity, we use the symbols $e$
and $t$ for both the entity/relation itself and for its
vector.

\devour{
\footnote{While
typically the count of unique entities is significantly
lower than the number of relations, the quantity of
entities can still reach up to millions if the framework
encounters a vast corpus of documents. Hence, we employ
Hierarchical Navigable Small World graphs
(HNSW) \citep{malkov2018efficient} to facilitate rapid
indexing for an extensive number of entities.}
}
\vspace{0.95em}
Our \textbf{query format}
for querying the memory is:\mathlinebreak
\makebox[\linewidth]{
$\mathbf{q} \in \{\langle e^q_s, t^q, *\rangle,\langle *, t^q, e^q_o\rangle\}$
}   \mathlinebreak
where $e^q_s$, $e^q_o$, $t^q$ are subject entity, object entity, and
relation. $*$ indicates the position in
the triple of
the entity we are querying for.
These two templates give us sufficiently
specific queries (as opposed to, e.g., queries with two variables)
that are likely to return useful entity information.

We want to retrieve triples from the memory that match the
query. 
Given that
the surface form of an entity (and also the relation) can vary (e.g., ``US'' vs
``USA''),
our match criterion is not exact match, but rather vector
similarity. We refer to entities/relations that are similar
to the query entity and the query relation
as \emph{candidate} entities/relations. 

For retrieval, we first determine a set of candidate
entities:\mathlinebreak
\makebox[\linewidth]{
${\cal C} = \{e'|\cos(e^q,e')\geq \tau_e \}$}\mathlinebreak
That is, all
entities with an above-threshold similarity are considered
candidate entities.

Similarly, we determine a set of candidate relations:\mathlinebreak
\makebox[\linewidth]{${\cal T} = \{t'|\cos(t^q,t')\geq \tau_t \}$}

If the query is a query for an object, i.e., 
$\mathbf{q} = \langle e^q_s, t^q, *\rangle$, then we
retrieve the following final set
${\cal E}$ of entities from the memory:
\[\{e_o|
\exists (e,t,e_o) \in
{\cal M}:e \in {\cal C},
t \in {\cal T},
0.5(\cos(e,e^q_s)+\cos(t,t^q))\geq \tau_r\}\]
where ${\cal M}$ is the memory. That is, we retrieve all
triples with entities/relations from the
candidate sets such that their average similarity to query
subject and relation is above the threshold.
Subject queries are handled analogously.\footnote{We discuss how we set the thresholds and other hyperparameters in Appendix \ref{sec:apdx_ft_details}.}

\subsection{Memory-API and Inference}
We now describe the
API that specifies how the LLM initiates memory writes and
memory reads.

\devour{\footnote{For sentence splitting,
we use NLTK \citep{bird-loper-2004-nltk}.}}

\begin{figure*}[t]
    \centering
    \begin{subfigure}[b]{0.96\textwidth}
        \centering
        \includegraphics[width=0.96\linewidth, trim=0 0 0 0, clip] {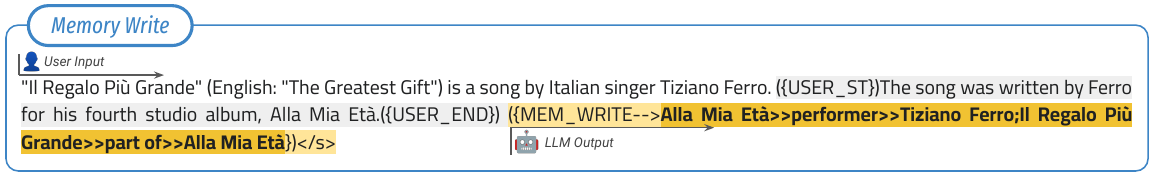}
        \caption{For memory writes, the input is given in
          two parts. (i) The pretext  provides context for
          the model (e.g., antecedents for pronouns). (ii)
          The \hlc[my_gray]{focus sentence}
          is the span of text
(bracketed by \texttt{(\{USER\_ST\})} and \texttt{(\{USER\_END\})})
          from which the model is tasked to
extract all relations.
The model 
calls the API starting with the
\hlc[my_lyellow]{\texttt{(\{MEM\_WRITE-{}->}} command
followed by the \hlc[my_dyellow]{extracted relations}.
\hlc[my_lyellow]{\texttt{\})</s>}} closes the
API call.
In each  document,
\framework scans the sentences one by one.}
        \label{fig:memory_write_inference}
     \end{subfigure}\vspace{5pt}
     \begin{subfigure}[b]{0.96\textwidth}
        \centering
        \includegraphics[width=0.96\linewidth, trim=2 1 2 1, clip] {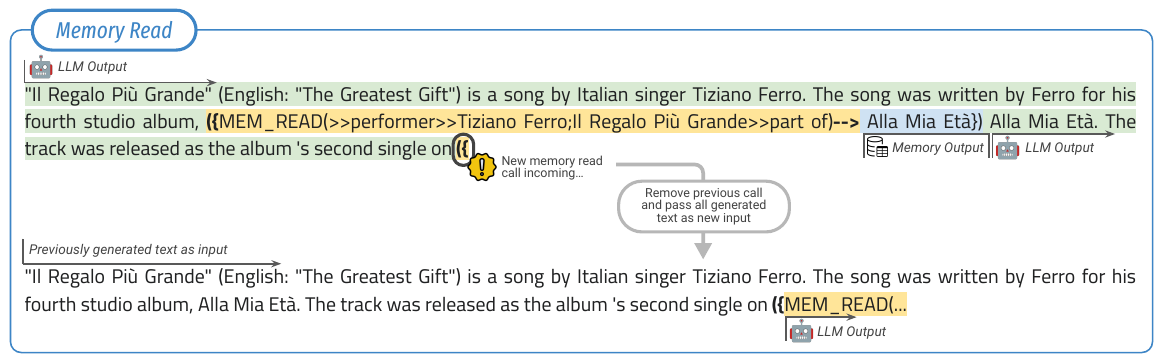}
        \caption{The model
decodes one token at a time, as in standard
language modeling. It is also trained to generate
memory read commands at points when
they can retrieve useful
information.
In the example, after decoding \hlc[my_green]{some
  tokens}, the model generates a
\hlc[my_lyellow]{\texttt{(\{MEM\_READ(}} command followed by
\hlc[my_lyellow]{queries}. 
\hlc[my_lyellow]{\texttt{-{}->}}  triggers execution of
the queries. \hlc[my_blue]{Returned results} are appended.
        The model then uses the retrieved results for 
         decoding \hlc[my_green]{the
           posttext}.
         Whenever, during further decoding, the model
         initiates a new memory read by emitting
        \hlc[my_lyellow]{\texttt{(\{}},
 we remove
the previous one
because it is unlikely to still be useful.}
        \label{fig:memory_read_inference}
     \end{subfigure}
    \caption{MemLLM inference with memory read and memory
      write}
    \label{fig:framework_workflow}
\end{figure*}

\textbf{Memory writes.}
\label{sec:memory_write_inference}
We process the input sentences one  by one.
See the example given in
Figure \ref{fig:memory_write_inference}.
For sentence $s_i$
the input $x^\text{MW}_i$ to the LLM is formatted as follows:\mathlinebreak
\makebox[\linewidth]{$x^\text{MW}_i\! =\! S_{<i}\! +\! \texttt{(\{USER\_ST\})}\! +\! s_i\! + \!\texttt{(\{USER\_END\})}$}\mathlinebreak
where $S_{<i}$ are the $i-1$ preceding sentences and $s_i$
is bracketed by tags to mark it as
the \emph{focus sentence}.
The LLM's task is then to extract all relations occurring in
the focus sentence and to generate a write command
that
stores them in the memory:\mathlinebreak
\small
  \makebox[\linewidth]{$y[x^\text{MW}_i]=\texttt{(\{MEM\_WRITE-{}->}e_s^1\texttt{>>}t^1\texttt{>>}e_o^1;e_s^2\texttt{>>}t^2\texttt{>>}e_o^2;\ldots\texttt{\})}$}\mathlinebreak\normalsize
Context $S_{<i}$ is necessary to
extract relations from the focus sentence, e.g.,
if the focus sentence refers to a previously introduced
entity with a pronoun. We finetune the LLM to only extract
relations from the focus sentence (not from the preceding
context); see \S\ref{sec:finetuning} for details.

To extract all relations from a document and write them to
the memory, we iterate over
the sentences of a document one by one.

% \enote{hs}{can you please provide the text that you
%   screenshot in the figures?}

%\enote{hs}{I don't think you give an example for a memread
%  command? you should show the entire process, which may
%  take four different snapshots of the model's input/output,
%  i.e., two snapshots in addition to the two you present}

\devour{
\footnote{We decode with a late stopping technique,
not with greedy decoding. See Appendix \ref{sec:apdx_memwrite_decoding}.}}

\textbf{Memory reads.}
\label{sec:memory_read_inference}
The LLM can at each point in time
either emit a regular token or 
initiate an API MEM\_READ call by generating:\mathlinebreak
\mathindent $\texttt{(\{MEM\_READ(}$\mathlinebreak
It then continues by generating 
subject or object queries as introduced above:
$\mathbf{q} \in \{\langle e^q_s, t^q, *\rangle,\langle *,
t^q, e^q_o\rangle\}$. The syntax for
the memory-read API call is:\mathlinebreak
\mathindent $\texttt{(\{MEM\_READ}(e^{q1}_s\texttt{>>}t^{q1}\texttt{>>};\texttt{>>}t^{q2}\texttt{>>}e^{q2}_o;\ldots)\texttt{-{}->}$\mathlinebreak
The entity sets $\cal E$ are then retrieved from the memory 
(\S\ref{sec:mem_struct}), merged  and appended to
the API call:\mathlinebreak
\mathindent$\texttt{(\{MEM\_READ}(e^{q1}_s\texttt{>>}t^{q1}\texttt{>>};\ldots)\texttt{-{}->}e_1,e_2,e_3,\ldots\texttt{\})}$\mathlinebreak
The LLM then continues decoding.
Figure \ref{fig:memory_read_inference} gives an example. The
LLM starts generating a sentence that refers to an album
by the Italian singer Tiziano Ferro. It has learned
that just before naming the album is a good point at which
to initiate a memory read.
Two queries are generated (including: ``What is the song 
``Il Regalo Più Grande'' part of?''). One entity is returned
by the memory (``$\texttt{-{}->}$ Alla Mia Età\})'') and written to the buffer.
The LLM then generates the name of the correct album (``Alla Mia Età.''). This
example illustrates that our memory has the potential of
reducing hallucinations because
through the memory
an explicit representation
is available
of the fact that ``Il Regalo Più Grande'' is part of the
album ``Alla Mia Età''.

\devour{
\footnote{We continue with the
second highest scoring token instead
of \texttt{(\{MEM\_READ(} if this happens.}
}

We remove memory-read API calls from the context if they
are no longer useful.
This happens in three cases:
(i) The returned set
$\cal E$ of entities is empty.
(ii)
The number of retrieved
entities
exceeds a threshold $Q_{thr}$ ($Q_{thr} = 30$).
%It is difficult for the
%model to identify the correct entity from a large set, so
Such large retrieval results are unlikely to be helpful.
(iii) The model
emits ``\texttt{(\{}'',
initiating a new memory-read API call.

For (iii), our motivation for removing the
API call is as follows.
Omitting API verbiage
preserves the text's natural flow and reduces the context to
those parts of the input that are still informative for
high-quality generation. 

\begin{figure*}[t]
    \centering
    {\includegraphics[width=0.9\linewidth, trim=1 1 1 1, clip] {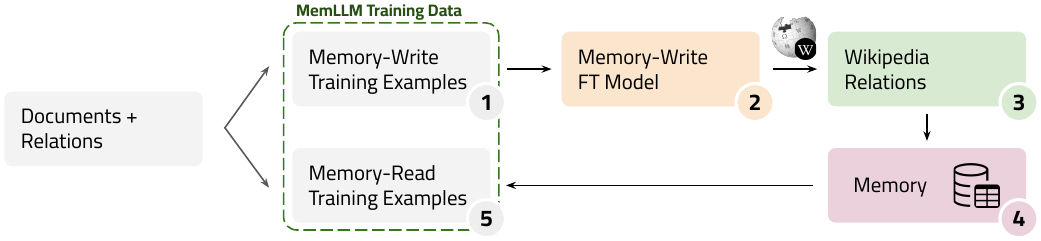}}
    \caption{\framework training data pipeline.}
    \label{fig:framework_data_pipeline}
\end{figure*}

\subsection{Finetuning the LLM}
\label{sec:finetuning}
We now describe how we create the dataset for
finetuning the model to generate memory-write and
memory-read API calls.  One innovation of our work is that
we create these API training data from
corpora
annotated with entities and
relations,
including
Re-\docred 
\citep{tan-etal-2022-revisiting},
a Wikipedia corpus annotated in Wikidata format with named
entity mentions and relations
(see also \S\ref{sec:setup})
that we will use as an
example below.

\textbf{Memory-write data.}
Figure \ref{fig:framework_data_pipeline} shows how 
we use Re-\docred's  annotated
relations. 
For each sentence $s_i$, we
retrieve from Re-\docred all relation triples such that 
one entity has a full mention (i.e., not a pronoun) in $s_i$
and the other entity has a full mention either in $s_i$ or in
the pretext ($S_{<i}$).
The memory-write training example  consists of 
the context $x^\text{MW}_i$ and the memory-write command
$y[x^\text{MW}_i]$;
see 
\S\ref{sec:memory_write_inference} and
Figure \ref{fig:memory_write_inference}.
Since we want to teach the LLM to generate
memory-write API calls, we compute the training loss on
$y[x^\text{MW}_i]$ only.

The set of
relation triples can be empty for a sentence $s_i$. In that
case we generate a  memory-write
command in $y[x^\text{MW}_i]$
that contains no relations. This encourages  the LLM
not to generate spurious relations for such ``empty'' sentences.

\textbf{Memory-read data.}
For effective memory reads, the LLM has to learn (i) to
identify where to initiate a query to the memory, (ii) to
generate queries that retrieve helpful information from the
memory and (iii) to make good use of the information that is
returned by the memory. We now describe how we
generate our training data with 
all three capabilities in mind.

% We generate multiple training instances $d'$ from a docred
% document $d$. To process $d$, we scan its annotated entity
% mentions from the beginning to the end of the document. For
% each entity mention $e_1$, we collect all relation triples
% in which it participates. Of these triples, we keep only
% those in which the mention of the second entity $e_2$ that
% the triple refers to has already occurred.  (The LLM will in
% general not be able to generate a query containing $e_2$ if
% $e_2$ has not yet occurred.)  

Given a
Re-\docred document $d$,
we generate a  different training
instance $d'$ for each memory-read API call.
To produce $d'$, we 
scan $d$'s annotated entity mentions from the beginning to the
end of the document. For each entity mention $e_\text{target}$,
we collect all relation triples in which it
participates. Such triples are a good basis for
memory-read API calls that -- when issued before
$e_\text{target}$ first appears --  will help the LLM to
correctly generate $e_\text{target}$; this is why we refer
to $e_\text{target}$ as the target entity.
We keep only that subset of the triples in which the mention of
the other entity $e_q$ that the triple refers to (the query entity) has already occurred.
(The LLM will in
general not be able to generate a query containing $e_q$ if
$e_q$ has not yet occurred.)
We also discard all triples
that we previously encountered during our scan.
(These are
already known at this point, so there is little utility
initiating a query for them.)  We then generate a
query for each remaining triple: either $\langle e_q, t,
*\rangle$ ($e_\text{target}$ = object) or 
$\langle *, t, e_q\rangle$ ($e_\text{target}$ = subject).  The memory-read API call for the queries generated
for $e_\text{target}$ is placed immediately preceding $e_\text{target}$. This will
retrieve $e_\text{target}$ from the memory in many cases and then make
it easy for the LLM to correctly predict $e_\text{target}$ at the next position.

Next we retrieve results for the query from the memory
(see Figure \ref{fig:memory_schema}).
The
memory we use here is the one that is populated from
Wikipedia by the trained memory-write model (as described
earlier in this section and 
in Figure \ref{fig:memory_write_inference}).  The memory write model misses some relations and
incorrectly identifies others, resulting in an imperfect
memory.  We intentionally use this imperfect memory because
it aligns the training data with the ultimate
inference conditions.

%This simulates
%our intended scenario of longterm and large scale use of the
%memory.  

If the query returns a large
number of results from memory (more than $Q_{thr} = 30$), we discard it
as unlikely to be helpful.
(See
Appendix \ref{sec:apdx_filt_amb_queries} for details.)
An example is the query $\langle
*, \text{country}, \text{United States}\rangle$ where Wikidata
defines the relation ``country'' as ``sovereign state that
this item is in''.
There are 
thousands of entities that satisfy this query. Such an
unspecific result is not useful.
Otherwise we add the queries and the query
result to $d'$; see 
Figure~\ref{fig:memory_read_inference} and
\S\ref{sec:memory_read_inference}.

Finally, we add the rest of $d$ to 
$d'$ up
to the next memory read (indicated by
``(\{'') or (if there isn't one) the entire remainder of $d$.

\devour{
If a subsequent training example ends up with the exact same
queries as the preceding one (which returns the exact query
results), we merge the posttext of the previous example with
the subsequent one and delay the next memory read
initiation. This means that the \framework would be trained
to make less redundant queries back-to-back and use the
query results for the post text as much as possible.
}

To summarize, each training example $d'$ is a concatenation of
(i) the pretext, including the first two letters
(``\texttt{(\{}'')
of the API
call, (ii) the API call proper
``\texttt{MEM\_READ(}$e^{q1}_s\texttt{>>}t^{q1}\texttt{>>};\ldots\texttt{)-{}->}$'',
(iii) the query result from the memory
``$e_1,e_2,e_3,\ldots\texttt{\})}$'' and (iv) the
posttext. The posttext consists of the rest of the
following text until the next memory read
or (if there
isn't one) the entire
rest of $d$.

%; in
%that case the posttext ends with ``\texttt{(\{}''.

The loss is applied to the API call (ii) -- this teaches the
model to generate the correct API call. The loss is also
applied to the posttext -- this teaches the model (a) to make
good use of the information provided in the query result
for predicting entities and (b) to
predict the next memory read (as indicated by
``\texttt{(\{}'').  (iii) is not subject to the loss since
the query results are generated by the memory, not by the
LLM.  For the training example $d'$ that contains the very
first ``\texttt{(\{MEM\_READ(}'' in the document (and only
for this $d'$), the loss is
also applied to the pretext -- because the LLM needs to
learn where to generate this first ``\texttt{(\{MEM\_READ(}''.
For a more comprehensive overview of the memory-read data generation process, refer to Appendix \ref{sec:apdx_algo_mr_data_generation}, which outlines the detailed algorithm.

\section{Experiments}
\label{sec:experiments}
\subsection{Setup}
\label{sec:setup}
To train and evaluate \framework, we construct training and
evaluation datasets as described in
\S\ref{sec:finetuning}. We require datasets annotated with
entities and relations.  We use three such datasets.  (i)
Re-\docred \citep{tan-etal-2022-revisiting}: Wikipedia texts
annotated (in a Wikidata format) with named entity mentions,
coreference information and 96 relations (occurring intra-
and inter-sentence).  Re-\docred includes many relation
instances missing in \docred \cite{yao-etal-2019-docred}.
(ii) \docred's distant supervised training set. It includes
$>$100K documents but fewer relations per document.  The
size of this dataset makes the training more effective and
robust.  (iii) A set of ``counterfactual'' variations of
Re-\docred
\citep{modarressi-etal-2024-consistent}. \citet{modarressi-etal-2024-consistent}
introduces an entity replacement strategy to find and apply
suitable replacements over Re-\docred.  In our initial
tests, we found that teaching the model to produce
counterfactual answers (which often contradict its
parametric memory) increases robustness against pretrained
knowledge bias. This is described in detail in \S\ref{sec:filt_rels}.

We finetune two Mistral-7B \citep{jiang2023mistral} models
using LoRA \citep{hu2022lora}, a memory-write model and a
memory-read model.  See Appendix \ref{sec:apdx_ft_details}
for details on finetuning and hyperparameters. Our baselines
are the original Mistral-7B and the memory-read model with
its memory capabilities disabled.  The latter baseline lets
us ascertain to what extent improvements are due to
in-domain finetuning (as opposed to the memory).

\subsubsection{Filtering Distant Supervision Relations}
\label{sec:filt_rels}
Re-\docred is 
human-annotated and mostly consists of relations with explicit evidence.
In contrast, the distant supervised \docred training set
lacks explicit evidence and contains many false positives
due to its automated annotation
method. To address this, we implement a few-shot-based
filtering approach to remove false-positive
relations.  We also apply this filtering to
Re-\docred relations that lack explicit evidence.

To increase the number of training examples, we also include examples from the distant supervision subset of \docred. Distant supervision \citep{mintz-etal-2009-distant} assumes that a relation $r$ exists between two entities ($e_s$, $e_o$) in a text if the text includes both entities and the $r = \langle e_s, t, e_o \rangle$ relation triple exists in a knowledge base. While this method is valuable for relation extraction, it may introduce noisy examples without any evidence of the relation in the text. This noise could adversely affect our training pipeline.

The experimental setup is as follows: We start with a partial document ($S = \{s_1,s_2,\ldots ,s_i\}$) mentioning two entities ($e_1$, $e_2$), with at least one of them present in the last sentence (i.e., the focus sentence), $s_i$. Our aim is to determine whether the potential relation $r$ between $e_1$ and $e_2$ has any evidence in the last sentence.

To filter out negative examples, we use large language
models (i.e., Mixtral). We design 8-shot in-context learning
examples to detect if there is evidence of a relation in the
focus sentence. We curate a test set to evaluate
the performance of this filtering mechanism as follows. We select 1000
examples from the human-annotated split of \docred as
positive examples where the focus sentence is annotated as
evidence. For negative examples, we choose 1000 examples
where the focus sentence contains at least one entity but
there is no evidence for the relation in the focus sentence.
\begin{table}[t]
	\setlength\tabcolsep{4.6pt}
	\centering
	\begin{tabular}{lrrrr}
		\toprule
		\textbf{Filtering Approach} & \textbf{Prec.} &\textbf{Rec.} & \textbf{F1} & \textbf{Acc.} \\
		\midrule
		Baseline & 0.58 & 0.83 & 0.68 & 0.61 \\
		Justification & 0.56 & 0.82 & 0.66 & 0.59 \\
		Reasoning & \textbf{0.78} & \textbf{0.84} & \textbf{0.80} & \textbf{0.80} \\
		\bottomrule
	\end{tabular}
	\caption{Comparing performance of different prompting strategies for filtering distant supervision data. The reasoning approach similar to chain-of-thought prompting performs best among the strategies.}
	\label{tab:filtering_performance}
\end{table}

For prompting, we apply three different strategies. In the first approach (baseline), we expect the LLM to answer with ``Yes'' or ``No'' to report whether the focus sentence contains evidence. With the second approach (justification), we expect the LLM to provide justification after giving the answer. In the final approach (reasoning), we expect the LLM to generate a natural sentence representing the relation, then provide reasoning, and finally generate the answer with ``Yes'' or ``No'' in the last sentence, similar to chain-of-thought prompting.

We present the results in Table \ref{tab:filtering_performance}. These results suggest that the reasoning approach outperforms the other two approaches by a large margin. Also, it suggests that the filtering would lead to higher quality based on the high recall score, 0.84. We demonstrate the best-performing prompt in Appendix \ref{sec:apdx_filt_prompt}.

After applying this method, we use the filtered distant dataset alongside 10 counterfactual variations of Re-\docred to generate data for the initial fine-tuning phase. We then continue the finetuning process with data generated from the supervised set of Re-\docred.
% EDITEND

\subsection{Perplexity evaluation}
To evaluate how the memory component would improve language modeling, we perform a perplexity evaluation. For this, we need a corpus to extract facts, do memory writes, and store them in structured memory. Our primary source is a full dump of English Wikipedia\footnote{Dump date: 2023-11-01, available at: \url{https://huggingface.co/datasets/wikimedia/wikipedia}}, but we also evaluate using Wikipedia abstracts and Re-DocRED texts for further analyses. The language modeling evaluation then examines how the model performs in terms of perplexity once the memory component has been filled.
Following \citet{liu-etal-2022-relational}, we report: (1)
OVERALL PPL (PPL on the entire input text), (2) TARGET PPL
(PPL on the target entities) and (3) ENTITY PPL (PPL on all
named entities). The model produces a token $w_i$ with
probability    $p(w_i|w_{<i})$:\mathlinebreak
\makebox[\linewidth]{$p(w_i|w_{<i})=p(w_i|w_{<i},\text{MR})p(\text{MR}|w_{<i})$}\mathlinebreak
\makebox[\linewidth]{$\phantom{p(w_i|w_{<i})=}+p(w_i|w_{<i})(1-p(\text{MR}|w_{<i}))$}\mathlinebreak
where $p(\text{MR}|w_{<i})$ is the probability of initiating
a memory read (MR) with the
``\texttt{(\{}'' token.\footnote{We evaluate $p(w_i|w_{<i})$ by setting $p(w_i|w_{<i},\text{MR})$ to zero for all positions except for positions where memory reads actually occur.
% We evaluate $p$ by setting
% $1-p(\text{MR}|w_{<i})$ to
% zero for all positions except for positions where memory
% reads actually occur.
% (Of course, we do not set $p(\text{MR}|w_{<i})$ to 1!)
The reason is that
taking into account
an MR call at each position results in a tree with
$2^n$ leaves at position $n$ in the text, each requiring a
memory call. This is too expensive to compute. As a result,
we evaluate with smaller values of $p(w_i|w_{<i})$ than
the true $p(w_i|w_{<i})$ estimated by the model and,
consequently, with higher perplexities, thus unfairly
penalizing \framework. Note that this is a problem for
fairness of our
perplexity evaluation, but not for a
real application (where
we only pursue a single path at each point).}

\begin{table*}[t]
	\small
	\setlength\tabcolsep{5pt}
	\centering
	\begin{tabular}{l@{\hspace{20pt}}|@{\hspace{10pt}}c@{\hspace{10pt}}|ccc}
		\toprule
         & \multirow{2}{*}{\textbf{Memory}} & \multicolumn{3}{c}{\textbf{PPL}} \\
		 && \textbf{OVERALL} &\textbf{TARGET} & \textbf{ENTITY} \\
		\midrule
        Baseline \#1 (Mistral-7b) &\multirow{2}{*}{(no memory)}& 5.823 & 3.550 & 4.666 \\
        Baseline \#2 (Memory Disabled)&& 4.997 & 3.510 & 4.353 \\
        \midrule
        \circled{1} \textbf{\framework} &MW[\emph{Wikipedia (Full)}] & 4.905 & 2.986 & 4.187 \\
		% \qquad + Gold MR Position & 1.602 & 1.007 & 1.316 \\
        % \qquad + Gold Queries & 1.598 & 0.901 & 1.287 \\
        % \qquad + Gold Target & 1.535 & 0.337 & 1.124 \\
        \midrule
        \circled{2} \framework & MW[\emph{Wikipedia (Abs.)}] & 4.898 & 2.955 & 4.170 \\
        % \midrule
        % \multicolumn{4}{c}{Memory: MW[\emph{DOCRED(Val.)}]}\vspace{3pt} \\
        % \framework & 1.594 & 0.963 & 1.300 \\
        \midrule
        \circled{3} \framework & \multirow{2}{*}{MW[\emph{Re-\docred (Test)}]} & 4.863 & 2.821 & 4.102 \\
        \qquad \circled{4} + Gold MR Pos. \& Queries && 4.634 & 1.938 & 3.548 \\
        \midrule
        \circled{5} \framework &\multirow{5}{*}{Re-\docred (Test)}& 4.811 & 2.596 & 3.993 \\
        \qquad \circled{6} + Gold MR Position && 4.674 & 2.232 & 3.728 \\
        \qquad \circled{7} + Gold Queries && 4.431 & 1.364 & 3.149 \\
        \qquad \circled{8} + Gold Target && 4.426 & 1.357 & 3.142 \\
        \qquad \circled{9} + Gold Target Only && 4.385 & 1.194 & 3.026 \\
        % \qquad + Gold MR Pos. \& Queries & 1.562 & 0.619 & 1.202 \\
    % \midrule
    % \multicolumn{5}{c}{\emph{Without late stopping}}\vspace{3pt} \\
		% Distant. & 0.541 & 0.325 & 0.587 & 0.563 \\
    % \qquad w/ Filtering + 85\% NoRel. Drop + 10xAnnot. & 0.562 & \textbf{0.448} & \textbf{0.818} & 0.666 \\
		\bottomrule
	\end{tabular}
	\caption{\framework performance on OVERALL PPL
          (all text), TARGET PPL (target
          entities) and ENTITY
          PPL (all entities).
We show  
          the effect of  memory content (``\textbf{Memory}'').
``MW[\emph{X}]'': the memory is
          populated with triples generated by
          memory-writes with \framework run on
          \emph{X}. \circled{5}--\circled{9}: the triples
          are from
Re-\docred (Test),
          Re-\docred's validation set.}
	\label{tab:memory_read_PPLs}
\end{table*}

In case of MR, $w_i$ is conditioned
on both MR
(including the MR call and the returned result, see
Figure~\ref{fig:memory_read_inference})
and the pretext $w_{<i}$. If there is no MR, then
$w_i$ is only conditioned on $w_{<i}$.

\devour{
Since perplexity is a measure based on a
given text as a gold output, gold memory-reads need to be
determined based on the text.  We follow our method on
creating the memory-read training data and use the memory
calls as the gold memory-read (MR) positions as the pre-determined
positions. }

% \footnote{}

%As the
%memory calls are produced mid-generation and would require a
%generation process to produce queries, we adhere to the
%inference process introduced
%in \S\ref{sec:memory_read_inference} and measure perplexity
%results by evaluating the CE loss upon generating each token
%of the text.

%Afterwards, we assess the principal objective of \framework,
%specifically evaluating its efficacy in language modeling
%while using its memory capabilities. We evaluate the
%performance of the framework in various conditions and

\label{sec:results}
Table \ref{tab:memory_read_PPLs} gives perplexity
results on Re-\docred test.
\framework outperforms the two baselines on
all three PPL measures (\circled{1}). This increase for
triples appearing for the first time in the text suggests
that memory-reads successfully recall relevant information
for language modeling. This improvement benefits not just
all entities in the text (ENTITY PPL) but the entire text
(OVERALL PPL).  TARGET PPL (the focus of this work) improves
by .524 (2.986 vs 3.510).
This substantial improvement demonstrates the effectiveness of \framework for
target entities. This capability is crucial for generating
factual text and preventing hallucinations.

For our \textbf{memory-read analysis,}
instead of using the LLM to write to the memory,
we directly
populate the memory with the relations from the
validation set (indicated as ``Re-\docred (Test)'' in
column ``Memory''). This lets us investigate what would happen if the
memory-write process were error-free, i.e., all remaining
errors are due to the memory-read process. We look at four
potential sources of error in memory reads and present in
each case an ablation in which this source of error is eliminated: the position of
the memory read is the gold position immediately before the
target entity
(\circled{6} ``Gold MR Position''),
the query to the memory
is the gold query 
(\circled{7} ``Gold Queries''),
the correct target entity is returned by the memory
(\circled{8} ``Gold Target''),
the correct target entity is returned by the memory and no
other entities
(\circled{9} ``Gold Target Only'').
\circled{9} is the lower bound perplexity for perfect memory reads (and
perfect memory writes). 

Comparing \circled{9} and \circled{8} on TARGET PPL  (1.194 vs 1.357)
shows the effect of ``ambiguity''. The +.343 gap is due to
the memory returning more targets than just the gold target.

Moving from \circled{8} to \circled{7} (1.357 to 1.364) indicates the impact of the memory retrieval process. 
In \circled{7}, we use the gold queries, but without ensuring the inclusion of the gold target in the results.
The next comparison 
highlights the impact observed when the LLM itself generates
queries
(\circled{6})
vs when only gold queries are issued
(\circled{7}).
Finally, the effect of the model itself selecting the position for
the memory read
(\circled{6})
versus predetermining that position (\circled{5}) is shown in the rise from 2.232 to 2.596.

% Since some queries are too general they could exceed
% $Q_{thr}$ (as discussed in Section \ref{sec:memory_read_inference}). 
% Therefore, the memory reads are omitted (due to the empty memory output) in this case in contrast to \circled{8} where it gets the target result as its only result. 

To isolate the factor
\textbf{memory-write performance},
we fix
(i) gold memory-read
positions and queries and (ii) the input corpus for extracting
relations (we use Re-\docred test). We only vary the
method by which the
memory is populated:
running the memory-write model on the input corpus
(\circled{4}) vs
reading out the relations from the gold data and directly
storing them in memory
(\circled{7}).
As expected, PPL
improves when directly stored (i.e.,  100\% precision and
recall)
triples are used (\circled{7}) vs when \framework
extracts and writes triples to memory (\circled{4}).
This indicates that there is room for improvement
by training \framework to do a better job at information extraction.

%  -- but
% not by much.
% It is worth noting that the recall of
% MW[docred(\emph{Val.})] is $0.578$ when assessed on the gold
% data from the validation set, i.e.,  $57.8\%$ of the gold
% validation set triples are correctly extracted and stored
% by the memory-write model. This indicates
% that \framework achieves good PPL results with
% memory-write extracted relations, despite their less-than-perfect
% coverage.
% The underlying reason is that specific target entities might be involved in multiple relation triples. Our memory-read calls can handle several queries simultaneously (as shown in Figure \ref{fig:memory_read_inference}). Thus, if at least one of these queries matches a relation triple that contains the target entity and returns it as the memory result, the memory-read call will be successful. This contributes to the effectiveness of the language modeling process despite the limited recall.

\textbf{Scaling the stored triples.}
In a real-world scenario, the size of the memory
and, consequently, the size of query results
will get large. This increases the risk
of unhelpful 
information being returned from the memory. To investigate this,
we compare our main
experiment (\circled{1}, using the full Wikipedia, 
111M triples) with two
ablations that use only Wikipedia abstracts (\circled{2},
38M triples)
and only Re-\docred test  (\circled{3}).
Table \ref{tab:memory_read_PPLs} shows that there is a relatively
small negative effect of memory size: \circled{3} (memory
stripped down to the relations generated from Re-\docred test) is only slightly better than \circled{1} (full
memory). This suggests good scaling properties of our approach.

\begin{table}[!t]
	\setlength\tabcolsep{5pt}
	\centering
    \begin{tabular}{l@{\hspace{20pt}}|cccc}
        \toprule
        \textbf{Method} & \textbf{REL} & \textbf{GEN} & \textbf{LOC} & \textbf{AVG}\\
        \midrule
        DEFER & 0.02 & 0.02 & 0.67 & 0.24 \\
        GRACE & \textbf{1.00} & 0.02 & \textbf{1.00} & 0.67 \\
        WISE & 0.70 & \textit{0.67} & \textbf{1.00} & \textit{0.79} \\
        \midrule
        \textbf{\framework} & \textit{0.78} & \textbf{0.76} & \textit{0.97} & \textbf{0.84} \\
        \bottomrule
    \end{tabular}
	\caption{Knowledge editing results on ZsRE with 1000
          sequential edits. AVG: mean of
 reliability (REL), generalization (GEN), locality (LOC).
Baseline results
(using the same model, Mistral-7B, and edit set) are
from \citet{wang2024wise}. Bold (italics): (second) best result.}
	\label{tab:model_editing}
\end{table}

\begin{wrapfigure}{r}{0.51\textwidth}
    \centering
    {\includegraphics[width=0.97\linewidth, trim=1 1 1 1, clip] {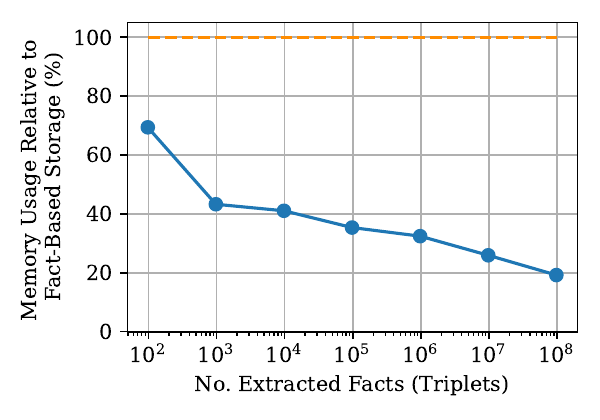}}
    % EDITSTART
    \caption{Memory efficiency of storing structured triples vs.\ proposition-based storage. The y-axis represents the fraction of memory required compared to a RAG system that stores an embedding per fact.}
    % EDITEND
    \label{fig:memory_redundancy}
\end{wrapfigure}

\textbf{Memory redundancy reduction benefits.}  Instead of
storing facts in a structured triple format, each fact could
be stored as its own proposition
\citep{chen-etal-2024-dense}. For instance, one can store
the triple $\langle\text{Washington D.C.}, \text{capital
  of}, \text{United States}\rangle$ as the proposition:
“Washington D.C. is the capital of the United States.” Based
on the memory populated with full Wikipedia (111M triples),
an equivalent RAG system would need to store and index 111M
proposition sentences. In a vector-based (dense) retrieval
setting, this means storing and indexing 111M vectors
alongside their text counterparts.  However, in \framework,
we store these facts in structured triples with identifiers
(cf. Section \ref{sec:mem_struct}), and the only
vector-based indices are entity and relation
embeddings. After storing the 111M triples, the entity/relation tables
contain roughly 21M unique entity/relation records.
%Because the relations
%contains 20.4M unique entity records.  Because the relations
%table contributes only a negligible number of
%vectors,
%\footnote{In rare cases, the model may extrapolate
%or generate relations outside the 96 used in training, but
%this is still under 5K and thus remains minimal compared to
%the main set of 96 relations.}
%Thus, the total number of vectors
%in \framework
%(entities plus relations)
%is less than
%one-fifth of what would be needed to index proposition
%sentences. 
Thus, as the number of extracted facts increases,
  our structured triple approach results in significantly
  lower memory usage and,
  at large scales, it requires less than 20\% of the memory
(21M)
  needed for direct proposition storage (111M), 
demonstrating its efficiency in reducing redundancy (Figure \ref{fig:memory_redundancy}).
Moreover, having only entities and relations encoded as vectors
reduces ambiguity and improves recall by introducing less
noise compared to encoding entire sentences.

\subsection{Knowledge Editing Evaluation}
To test whether
\framework facilitates knowledge editing, we
evaluate prompt-based knowledge editing. Following
\citet{hartvigsen2023aging} and \citet{editingLLMs2023Yao},
we measure  reliability (REL),
generalization (GEN) and locality (LOC). Each example
includes a prompt, an edit, a generalization test prompt and
a locality test prompt. The task is to apply the edit on the
original prompt to the model. The goal is for the model to respond
to original and generalization test prompts in accordance with the
edit. The locality test checks whether unrelated
knowledge is affected. An ideal method 
effectively applies edits, generalizes correctly and does
not harm unrelated knowledge.

Following \citet{wang2024wise}, we evaluate \framework on
ZsRE, a closed-book question answering
dataset \citep{levy-etal-2017-zero} with locality prompts
selected from Natural Questions
(NQ) \citep{kwiatkowski-etal-2019-natural}.  
We apply 1000 edits
from the evaluation set by
appending them to the end of the questions (the prompts)
using the following text: ``It is or they are'' and
bracketing them with tags. 
Example: ``\texttt{(\{USER\_ST\})}What city was Luca
Verdecchia born? It is or they are
Naples\texttt{(\{USER\_END\})}.'' The memory-write model
should then extract and store Verdecchia's place of birth, i.e., 
Naples.  We evaluate \framework
using a 5-shot QA prompt. The first four examples are
typical question-answer pairs. The fifth
in addition includes a full memory-read call. A prompt -- a generalization
or locality test prompt -- is appended to the 5 shots and 
a memory-read API call executed after the
question mark.

We
expect the model to answer the questions based on the
memory filled with the edits.  Some
edits in the dataset overlap or are intended to replace
previous edits. Therefore, if a newly
extracted triple has an exact matching entity and
relation  with an old triple, we replace the old one with
the new one.

%\subsubsection{Results}

Table \ref{tab:model_editing} compares knowledge
editing results for \framework with three baselines. 
\framework outperforms the baselines (AVG of .84).
High 
reliability (.78) and
generalization (.76) scores  suggest that \framework (i)
manages to extract and store the relation triple based on
the edit and (ii) utilizes the edit in the memory-read process to
answer the original and the generalization test questions
correctly. 
Moreover, since \framework uses an explicit
memory
the applied edits
only mildly affect the answers to unrelated questions:
\framework has a score of .97 on locality. This indicates that
there is little cumulative deterioration of the
explicit memory. 
% In contrast, many parametric memory approaches suffer from
% greater deterioration.

\paragraph{Qualitative Analysis.} 
Leveraging \framework's interpretable design, we identified
the causes behind the 22\% performance gap in reliability
from 1.00 to 0.78
(REL, \framework vs GRACE). Out of 216 errors, 45 were due
to  memory writes resulting in no triples or triples without the desired edit.
In 95 cases, the edit was captured in the memory write but
not retrieved by the memory read, either due to a bad query
or incorrect relation extraction during
the memory write.
Another 63 errors occurred when the model did not
effectively use the edits even though they were correctly
retrieved.
Many of these errors are due to 
the limitation to 96 relations
(see \S\ref{sec:experiments}). For example, the question
``How endangered does the IUCN consider Hyloxalus parcus?''
involves a relation that is not covered: ``IUCN conservation status''.
In another case, the question ``What family lineage was Xiao
Jia part of?'' retrieves the correct edit (``Southern Ming
Dynasty'') but for an incorrect
relation: \texttt{(\{MEM\_READ(Xiao Jia>>country of
citizenship>>)\})}, as the relation ``family'' is not one of
the covered 96. The model may then not recognize the query
result as relevant to the question and ignore it.
Addressing this limitation by supporting more relations  would resolve many of these errors. Even with this limitation and not being specifically designed for knowledge editing, \framework  outperforms other model editing methods. We believe that expanding its capability to handle a broader range of relations  would greatly enhance its performance. 

\section{Conclusion}
We present \framework, a novel
approach to endowing an LLM with
an explicit structured memory.
We publish  a training dataset that can be used to
extend any standard LLM with such a memory.
We show
that \framework improves language modeling (as
measured by entropy) and outperforms state-of-the-art
knowledge editing methods on ZsRE.

\devour{
cut: no space
The finetuning process trains the model to
initiate memory-write calls, allowing it to extract and
store relationships based on user input. Concurrently, for
the language modeling task, the model learns to initiate
memory-read calls during its decoding process:  it
generates queries and incorporates the memory's responses to
continue the language modeling (LM) task.  Based on the API,
we create a training dataset that can be easily used to
finetune any standard LLM to endow it with an explicit memory.  We show
that \framework improves overall language modeling
capabilities, significantly enhancing performance on texts
involving entities that make use of previously extracted
knowledge. Moreover, using a knowledge editing QA evaluation, we demonstrate that our method can outperform model editing methods in handling many edits at a time.
For future work, we intend to extend the
applicability of our solution across diverse domains by
incorporating additional types of relationships. 
}

\section*{Limitations}
While the structured relation-based memory improves factuality and interpretability, it has its own limitations. The current version of \framework handles only 96 relation types commonly used in Wikidata. However, to handle all types of knowledge extraction and storage, the model should be capable of extracting other types of relations. Composite relations that could be inferred from multiple already extracted relations are not detected or utilized in the current version of \framework. For instance, if we extract (California, country, United States) and (Apple Inc., located in, California), we expect the relation (Apple Inc., located in (or Country), United States) to be inferred. \framework is not a memory-aware solution. This means if a fact is not stored in the memory, but the decoding process generates a partial prompt that requires that fact, the model would either continue generation based on its parametric knowledge or hallucinate. 

We acknowledge that Retrieval-Augmented Generation (RAG) methods are widely used for a more grounded text generation. However, like other baselines in knowledge editing, we do not include RAG methods in our comparisons due to the difficulty of modifying facts in their unstructured format.
Updating a single fact would require locating and altering every related text snippet and embedding within the RAG’s knowledge base, which is highly impractical.
MemLLM's memory is not pre-populated as a KB; it stores edited facts through a memory-write process, similar to non- and semi-parametric editing methods. While methods like ChatDB \citep{hu2023chatdb} also employ non-parametric structured memory, they rely on database tables that are more suitable for analytical tasks. This setup requires the user to define a task-specific database schema in advance and explicitly prompt the model with it. In contrast, MemLLM uses a much more generalized memory structure that supports both language modeling and QA tasks (as discussed in the knowledge editing section) without requiring any changes to its approach or memory format. Furthermore, the nature of prompt-based approaches—compared to fine-tuning—makes them inherently less faithful and more dependent on the specific content stored in memory.

We refer all these limitations to future work, as in this paper we have laid the initial groundwork for building a more complex and comprehensive method.

\bibliography{tmlr}
\bibliographystyle{tmlr}

\appendix
\begin{table*}[t]
    \centering
    \begin{tabular}{lc}
        \toprule
        Query ($\mathbf{q}$) & Relation type ($t^q$) \\
        \midrule
        \multirow{8}{*}{$\langle *, t^q, e^q_o\rangle$}
        & country of citizenship, country, country of origin,\\
        &religion, place of birth, place of death, work location,\\
        &location, basin country, residence, location of formation,\\
        &publication date, production company, platform,\\
        &original language of work, applies to jurisdiction,\\
        &located in the administrative territorial entity,\\
        &headquarters location, inception,\\
        &employer, date of birth, date of death, educated at \\ & \\
        $\langle e^q_s, t^q, *\rangle$ & contains administrative territorial entity \\
        \bottomrule
	\end{tabular}
	\caption{List of ambiguous queries subject to the filtering process. }
	\label{tab:amb_queries}
\end{table*}
\section{Memory-write Decoding Method}
\label{sec:apdx_memwrite_decoding}
While one might use \framework with greedy decoding for memory writes, we suggest that the finetuned model may end the memory-write too early, before completely extracting all relations. Therefore, to ensure the model captures all relevant relations, we implement a late stopping strategy. In this approach, similar to greedy decoding, we consistently select the top-scoring token as the next token, unless it's the closing token "\texttt{)\}}". If the closing token scores highest, we note its position, calculate the average log probability score of the sequence up to that point, and proceed with the second highest scoring token—-typically the ";" separator—-resuming greedy decoding. By tracking the positions where the closing token was predicted, along with their corresponding logprob scores, we maintain the generation process until there are no enhancements in the scores for K=5 consecutive times. Subsequently, we halt the generation and select the position with the highest score as the cutoff point.

\section{Filtering Ambiguous Queries}
\label{sec:apdx_filt_amb_queries}
As we aim to assist the model with the stored memory content, having concise query results would facilitate reaching this objective. Getting precise outputs from the memory would require queries that are tailored in a way which lead to an exact match or related entities to the target entity. To reduce the chances of getting a vast and wide-range amount of outputs from the memory, we exclude queries that potentially leads to such results. In Table \ref{tab:amb_queries}, we demonstrate query patterns that we intuitively assume based on the queried entity and the relation type that would lead to an ambiguous result. Therefore, we drop any query that would match with one of the mentioned patterns.

\section{Memory-read Data Generation}
\label{sec:apdx_algo_mr_data_generation}
Algorithm \ref{algo:MR_data_generation} presents the pseudocode for the process of generating \framework’s memory-read training data. 
See Section \ref{sec:finetuning} for a detailed description
of the same process.

\begin{algorithm}[!ht]
  \footnotesize
  \caption{Memory-read Data Generation}
  \begin{algorithmic}[1]
    \Statex \textbf{Input:}
    \Statex \hspace{1em} $D$: Re-DocRED documents ($d \in D$)
    
    % \Statex \hspace{1em} $\tau_r$: 
    % \Statex \hspace{1em} $M_N$: Maximum number of alternatives to sample from
    \Statex \textbf{Output:}
    \Statex \hspace{1em} $\mathcal{D_\text{MR}}$: Memory-read training data ($d' \in \mathcal{D}$)
    \Statex \textbf{Auxiliary functions:}
    \Statex \hspace{1em} $\textsc{Triples}(e_\text{target}, d)$: Returns all triples in document $d$ containing the entity $e_{\text{target}}$ as the subject or object: \[\textsc{Triples}(e_\text{target}, d) = \{ \langle e_\text{target}, t, e_q\rangle \in d \} \cup \{ \langle e_q, t, e_\text{target}\rangle \in d \}\]
    \Statex \hspace{1em} $\textsc{QueryResult}(q, M)$: Returns the set of entities retrieved by query $q$ over memory $M$ (e.g. Wikipedia facts extracted using memory writes).
    \Statex \textbf{Definitions:}
    \Statex \hspace{1em} positionIdx: The position of an entity (or word) within the document text.
    \State Initialize SeenTriples $\gets \{\}$, SeenEntities $\gets \{\}$, $\mathcal{D}_\text{MR} \gets [\ ]$
    \For{$d$ in $D$}
        \State $\text{prevReadPos} \gets 0$, $\mathcal{D}_d \gets [\ ]$
        \For{$e_\text{target}$ in $d$, in order of appearance in text}
            \State $\mathbf{q} \gets \{\}$, $\mathbf{R} \gets [\ ]$
            \Comment{($\mathbf{R}$ is the query-results dictionary for the queries in $\mathbf{q}$)}
            \For{triple in $\textsc{Triples}(e_\text{target}, d)$}
                \If{triple $\notin$ SeenTriples}
                    \State Let $e_q$ be the other entity in $\text{triple}$
                    \If{$e_q \in$ SeenEntities}
                        \State $q \gets$ $\text{triple}-e_\text{target}$
                        \If{$q \notin \mathbf{q}$}
                            \State $R_q \gets \textsc{QueryResult}(q, M)$
                            \If{$|R_q| \leq Q_\text{thr}$}
                                \State Add $q$ to $\mathbf{q}$, $\mathbf{R}[q]\gets R_q$
                            \EndIf
                        \EndIf
                    \EndIf
                    \State Add triple in SeenTriples
                \EndIf
            \EndFor
            \If{$|\textbf{q}| > 0$}
                \State currentReadPos $\gets e_\text{target}.\text{positionIdx}$
                \State $d'.\text{pretext} \gets$ $d.\text{text[:currentReadPos]}$
                \State $d'.\text{queries} \gets \{\}$
                \State $d'.\text{results} \gets \{\}$
                \State Sort $\mathbf{q}$ in ascending order by $|\mathbf{R}[q]|$
                \For{$q$ in the first 3 elements of $\mathbf{q}$}
                    \State Add $q$ to $d'.\text{queries}$
                    \State $d'.\text{results} \gets d'.\text{results} \cup \textbf{R}[q]$ 
                \EndFor
                \If{$|d'.\text{results}| = 0$}
                    \State $d'.\text{results} \gets \{e_\text{target}\}$
                \EndIf
                \If{$|\mathcal{D}_d|>0$}
                    \State $\mathcal{D}_d[-1].\text{posttext} \gets d.\text{text[prevReadPos:currentReadPos]}$
                    \State $\text{prevReadPos} \gets \text{currentReadPos}$
                \EndIf
            \EndIf
            \State Add $e_\text{target}$ in SeenEntities
        \EndFor
        \State $\mathcal{D}_d[-1].\text{posttext} \gets d.\text{text}[\text{currentReadPos}:]$
        \State $\mathcal{D_\text{MR}} \gets \mathcal{D_\text{MR}} \,\Vert\, \mathcal{D}_d$ 
        \Comment{($\Vert$ denotes list concatenation)}
    \EndFor
    \State \Return $\mathcal{D}_\text{MR}$
  \end{algorithmic}
  \label{algo:MR_data_generation}
\end{algorithm}

\section{Hyperparameters Details}
\label{sec:apdx_ft_details}
We finetune \framework, with a Mistral-7B-v0.1 model \citep{jiang2023mistral} using an Adam optimizer \citep{KingBa15}, with the learning rate set to $2 \times 10^{-5}$, 2 epochs, and a batch size of 96. For LoRA specific parameters, we apply a dropout rate of 0.1, with a rank of 16 and an alpha weight of 8. 

We opted to set $Q_{thr}$ to 30 based on the distribution of triples observed in the Re-DocRED dataset. During the construction of memory-write data, we found that the 95th percentile of sentences contained a maximum of approximately 30 triples. This value serves as an upper threshold: if the number of entities exceeds 30, it is likely to surpass the typical number of triples within a sentence. Consequently, a higher count of entities could indicate that many of them are unrelated to the sentence's factual content, reducing their overall informativeness.

To select the memory retrieval hyperparameters (\S\ref{sec:mem_struct}), we must balance explicitness with the need to accommodate variations in entity mentions and relation types. This balance is influenced by the data and the entities involved, but generally, a larger $\tau_e$ increases explicitness. However, it also limits the number of similarly mentioned entities that can be retrieved, which depends on the use case. A smaller $\tau_e$ could retrieve more entities, but it would also increase query execution time.
The selection of $\tau_t$ depends on the supported relation types and the required flexibility in retrieving closely related relation types. For instance, in model editing, where handling loosely similar relation types is necessary, a more relaxed $\tau_t$ value is appropriate. Finally, $\tau_r$ determines the final number of outputs retrieved during the memory-read. A larger $\tau_r$ makes the memory more explicit in both entity and relation type.
We set $\tau_e$ and $\tau_t$ to 0.7 and $\tau_r$ to 0.85. We set these values to $\tau_e=0.85$, $\tau_t=0.2$ and $\tau_r=0.6$ respectively for model editing experiments.

\section{Filtering Prompt}
\label{sec:apdx_filt_prompt}
\begin{figure*}[t]
	\centering
	\includegraphics[width=0.93\linewidth]{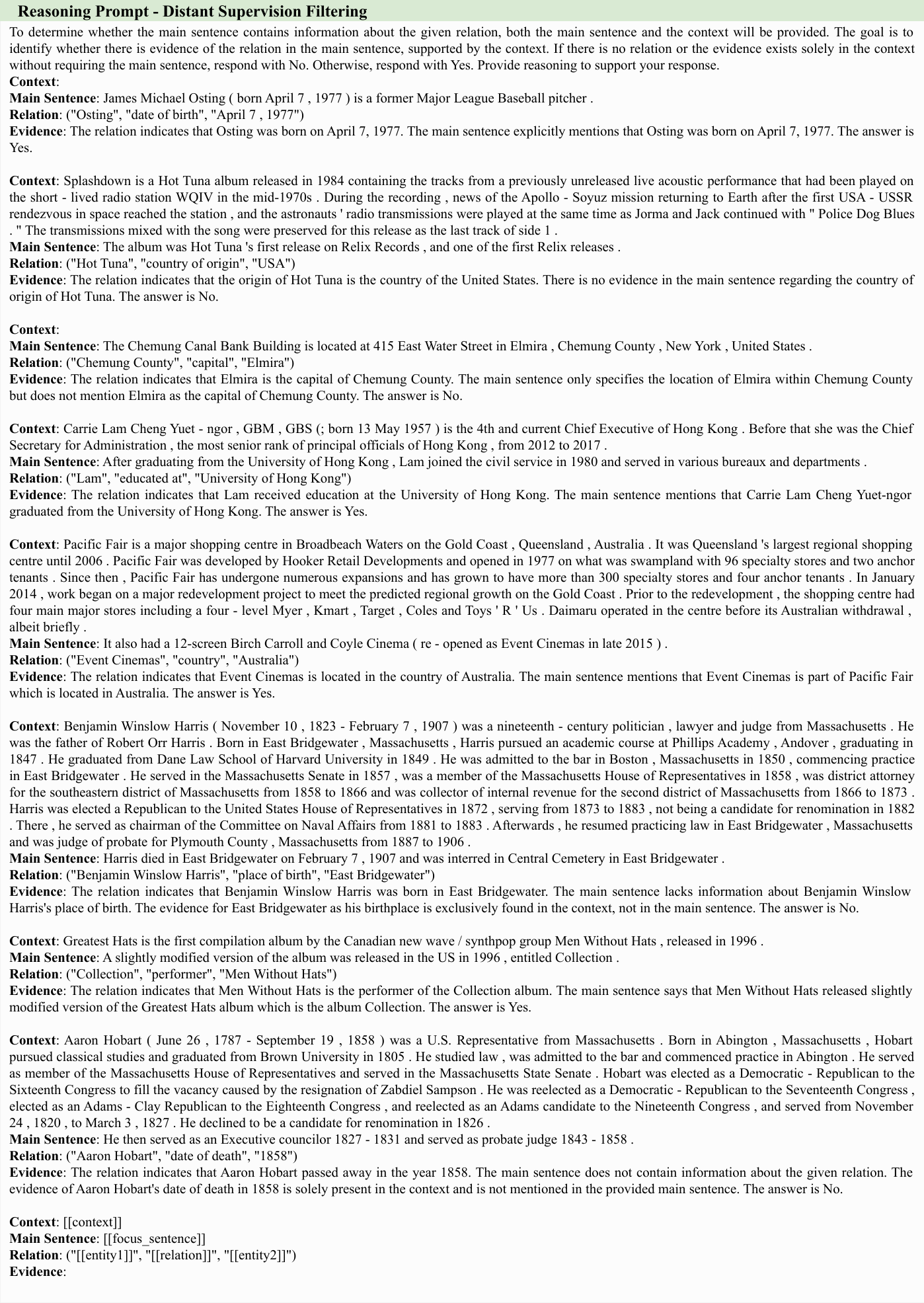}
	\caption{The prompt for the distant supervision dataset filtering. This prompt includes the natural representation of the relation, the reasoning, and the final answer.}
	\label{fig:distant_filtering_prompt}
\end{figure*}

In Figure \ref{fig:distant_filtering_prompt}, we demonstrate the best-performing prompt in our filtering process over the distant supervised subset of \docred.

\end{document}